\documentclass[letterpaper]{article} 
\usepackage{aaai24}  
\usepackage{times}  
\usepackage{helvet}  
\usepackage{courier}  
\usepackage[hyphens]{url}  
\usepackage{graphicx} 
\usepackage{natbib}  
\usepackage{caption} 
\usepackage{algorithm}
\usepackage{twemojis}
\usepackage{multirow}
\usepackage{bm}
\usepackage{booktabs}
\usepackage{amsmath}
\usepackage{times}
\usepackage{epsfig}
\usepackage{graphicx}
\usepackage{algpseudocode}
\usepackage{amssymb}
\usepackage{booktabs}
\usepackage{array}
\usepackage{makecell}
\usepackage{amssymb}
\usepackage{caption}
\usepackage{soul}
\usepackage{listings}
\usepackage{cuted}
\usepackage{capt-of}
\usepackage[super]{nth}
\usepackage{paralist,bbding,pifont}

\usepackage{newfloat}
\usepackage{listings}
\DeclareCaptionStyle{ruled}{labelfont=normalfont,labelsep=colon,strut=off} 
\floatstyle{ruled}
\newfloat{listing}{tb}{lst}{}
\floatname{listing}{Listing}
\title{CSL: Class-Agnostic Structure-Constrained Learning for \\ Segmentation Including the Unseen}
\author {
    Hao Zhang\textsuperscript{\rm 1},
    Fang Li\textsuperscript{\rm 1},
    Lu Qi\textsuperscript{\rm 2},
    Ming-Hsuan Yang\textsuperscript{\rm 2, \rm 3},
    Narendra Ahuja\textsuperscript{\rm 1}
}
\affiliations {
    \textsuperscript{\rm 1}University of Illinois at Urbana-Champaign\\
    \textsuperscript{\rm 2}University of California Merced\\
    \textsuperscript{\rm 3}Google Research\\
    \{haoz19, fangli3\}@illinois.edu, \{lqi5, mhyang\}@ucmerced.edu, n-ahuja@illinois.edu
}

\begin{document}

\maketitle

\begin{abstract}
Addressing Out-Of-Distribution (OOD) Segmentation and Zero-Shot Semantic Segmentation (ZS3) is challenging, necessitating segmenting unseen classes. Existing strategies adapt the class-agnostic Mask2Former (CA-M2F) tailored to specific tasks. However, these methods cater to singular tasks, demand training from scratch, and we demonstrate certain deficiencies in CA-M2F, which affect performance. We propose the Class-Agnostic Structure-Constrained Learning (CSL), a plug-in framework that can integrate with existing methods, thereby embedding structural constraints and achieving performance gain, including the unseen, specifically OOD, ZS3, and domain adaptation (DA) tasks. There are two schemes for CSL to integrate with existing methods (1) by distilling knowledge from a base teacher network, enforcing constraints across training and inference phrases, or (2) by leveraging established models to obtain per-pixel distributions without retraining, appending constraints during the inference phase. We propose soft assignment and mask split methodologies that enhance OOD object segmentation. Empirical evaluations demonstrate CSL's prowess in boosting the performance of existing algorithms spanning OOD segmentation, ZS3, and DA segmentation, consistently transcending the state-of-art across all three tasks. 
\end{abstract}

\section{Introduction}

Semantic segmentation is a fundamental task in computer vision, which associates with each pixel in a given image probabilities of belonging to different classes. Recent approaches have achieved remarkable results on several closed-set benchmarks that contain images from known classes, called In-Distribution (ID) images. 
However, the segmentation with the unseen, e.g., Out-Of-Distribution (OOD), Zero-shot-semantic (ZS3) segmentation, is always challenging because it requires segmentation and discrimination based on training on only ID images. 

The existing methods for such tasks can be distinguished by whether they use OOD data for training. Some methods expand the training set to include OOD images from other datasets ~\cite{oe, me, synboost, pebal}, or utilize large-scale models, e.g., SAM~\cite{sam} to generate region proposals. 
Such expansion-based approaches are not of great interest in this paper since we aim to solve the general problem of OOD segmentation without having access to any OOD images for training. 
We propose to learn models of objects that extend to classes beyond those in the ID set.

Existing segmentation methods typically infer OOD if some properties of the outputs are sufficiently different from those seen on ID images. 
An example of properties used is the uncertainty in pixel label prediction, as in the SML methods (Figure~\ref{fig1}). Other examples of properties used are errors in image reconstruction~\cite{imageresynthesis}, and the similarity to which results are perturbed by adversarial attacks ~\cite{liang2017enhancing, obsnet}. 
However, these methods result in noisy predictions due to a lack of structured knowledge. Current techniques, such as those in \cite{rba, ema}, tackle this issue using the region-based framework, Mask2Former (M2F)~\cite{mask2former}. However, to achieve optimal performance, they necessitate OOD data and complete train the model from scratch.
For ZS3 or open-word semantic segmentation, existing methods~\cite{zegformer, zsseg} typically leverage CA-M2F, trained on the ID set, as a region generator and utilize CLIP \cite{clip} to identify the semantic class for each region. Some works~\cite{qi2022open} empirically demonstrate that CA training benefits the performance on OOD data and since M2F decouples the per-pixel prediction task into $2$ sub-tasks: (1) mask prediction and (2) per-mask class prediction optimized by the mask loss and class loss, a straightforward way is removing the class loss and leverage hard assignment as post-processing during inference. However, our observations reveal that such adjustments are insufficient in eliminating class information. The process of hard assignment frequently results in unanticipated outcomes. For instance, certain objects might not have corresponding masks, and in certain situations, multiple objects may be erroneously blended into a single mask.

\begin{figure*}[!tp]
  \centering
  \includegraphics[width=.8\linewidth]{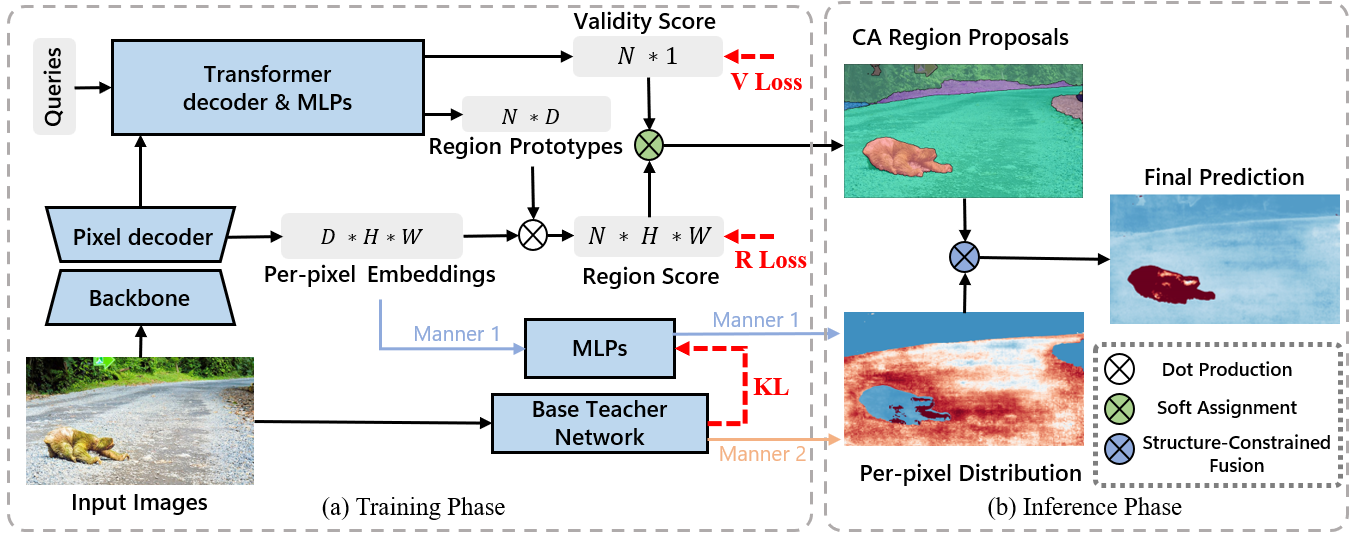}
  \vspace{0pt}
  \caption{Overview of our CSL framework. CSL consists of a backbone, a pixel decoder, a transformer decoder, a base teacher network, and MLPs. $N$ learnable region queries and the image features are fed to the transformer decoder and MLPs, to obtain $N$ pairs of latent region prototypes and their validity scores. 
  We calculate the normalized similarity between each element $\mathcal{E}_{h,w}$ of per-pixel embeddings and each $\mathbf{P}_n, n\in \{1,2,...N\}$ by simple dot production followed by the sigmoid function to get $N$ region scores. The validity scores indicate the degree of the region prototypes are valid for given images. During training, a valid loss, a region loss, and a distillation loss are used to optimize the model. Instead of assigning each pixel from the input image to one of the prior fixed classes, CSL assigns it to one of $N$ learnable region prototypes by our proposed soft assignment. During inference, we introduce structure-constrained Fusion to calculate the final prediction.}
  \vspace{0pt}
  \label{fig3}
\end{figure*}

In this paper, we present Class-Agnostic Structure-Constrained Learning (CSL) framework for seamless integration with existing methodologies, including OOD, ZS3, and DA segmentation, to improve their performance by incorporating structure constraints.
CSL offers two plug-in integration schemes:
(1) Knowledge distillation from a base teacher network, potentially any existing method, with structure constraints imposed during training and inference.
(2) Direct application of existing methods for per-pixel prediction, incorporating structure constraints solely during inference, bypassing retraining.
While the first style facilitates end-to-end training, the second negates the need for retraining, and both surprisingly yield comparable gains over foundational methods.
In semantic segmentation, annotations commonly amalgamate all instances of a class into a singular mask. We split this mask into multiple isolated components for training, mitigating bias from seen classes. During inference, CSL employs a soft assignment to derive region proposals at the disconnected-component level. Compared with the prevalent hard assignment, the soft assignment boosts the performance of unseen samples.
The main contributions of this paper are as follows:
\begin{itemize}
\item  We present CSL, a modular plug-in framework with 2 $2$ schemes, designed for seamless integration with established methodologies, enhancing the segmentation of unseen classes by incorporating structural constraints.

\item We propose mask split preprocessing, splitting class masks into isolated components, effectively attenuating the bias of seen class data. Furthermore, we employ a soft assignment in post-inference for region proposal generation and elucidate the driving factors behind the observed performance enhancements.

\item Through extensive experimental validation, we ascertain that CSL markedly enhances $10$ prevailing techniques across all three segmentation tasks, including OOD segmentation, ZS3, and DA segmentation, consistently outstripping state-of-the-art benchmarks.

\end{itemize}

\section{Related Work}

\subsection{Out-of-Distribution Segmentation}
\textbf{Uncertainty-based Methods}. Leveraging pixel-wise prediction uncertainty, OOD segmentation methods \cite{hendrycks2016baseline, lee2017training, liang2017enhancing, tian2021weakly, mcu} avoid retraining, thus saving computation. However, issues arise in hard-predicted regions. Jung et al. \cite{sml} refine boundary anomaly scores, while others \cite{kendall2017uncertainties, lakshminarayanan2017simple, mukhoti2018evaluating} apply MC dropout, often with limited success.
%
\textbf{Image Reconstruction}. Autoencoders and GANs dominate reconstruction methods \cite{baur2019deep, creusot2015real, di2021pixel,haldimann2019not, liu2020photoshopping}. Notably, ID-only trained models \cite{xia2020synthesize, lis2020detecting, ohgushi2020road, creusot2015real, imageresynthesis, jsrnet} effectively reconstruct ID samples, but falter with OODs, hindered further by domain sensitivity and extended training/inference times.
\textbf{Adversarial Attacks}. Adversarial attacks serve as OOD data simulators in image classification \cite{goodfellow2014explaining} and detection \cite{ma2018characterizing}. Besnier et al.'s ObsNet, though utilizing Local Adversarial Attacks, faces noisy prediction challenges due to structural information absence.
\textbf{Outlier Exposure}. The outlier exposure (OE) strategy by Hendrycks et al.~\cite{hendrycks2018deep} augments the training set with non-overlapping outliers. Conversely, some methods \cite{chan2021entropy, bevandic2019simultaneous, vandenhende2020revisiting, bevandic2018discriminative, rpl, pebal} embed OOD objects from datasets like COCO \cite{coco} and ADE20K \cite{ade20k}, potentially reducing OOD segmentation to mere binary segmentation due to overlaps.

\section{Proposed Method}

As shown in Figure~\ref{fig3}, CSL provides two schemes to plug in existing methods. The first is an end-to-end scheme, which distills the knowledge from the base teacher network to the CSL framework. The second scheme directly utilizes the existing models as a base teacher network to obtain per-pixel distributions and fuse them with the class-agnostic region proposals during inference without retraining them. Validity loss, region loss, and distill loss, which is the mean Huber loss between the predicted per-pixel distribution and the output of the base teacher network (only for scheme$_1$), are used for optimization.

\subsubsection{Class-Agnostic Training.} To capture the essential features of semantic classes that are applicable beyond the training classes, we design CSL in a class-agnostic way. It learns region prototypes characterized by visual appearance and spatial features and uses them to generate region proposals.
%
CSL firstly uses a backbone and a pixel decoder to generate multi-level feature embeddings $\mathcal{E}^{l}\in\mathbb{R}^{H_l \times W_l\times D_l}$, where $l \in \{4,8,16,32\}$ indicating the downsampled size of feature map compared to the original image. $D_l$ is the dimension of the embeddings. In addition, we have $N$ learnable queries, which cascadely interact with multi-level feature embeddings $\mathcal{E}^{l}$, where $l \in \{8,16,32\}$, to generate $N$ region prototypes $\mathbf{P}\in \mathbb{R}^{N\times 256}$. These prototypes act as centers for grouping the per-pixel embeddings $\mathcal{E} \in \mathbb{R}^{\frac{H}{4}\times \frac{W}{4}\times 256}$ and follows by a upsampling to get the region prediction $\mathbf{R} \in [0,1]^{N\times H\times W}$. The region scores $\mathbf{R}$, and the validity scores $\mathbf{V} \in [0,1]^{N}$ are fed into the soft assignment module (Sec 3.3) and generate region proposals.

\subsubsection{Comparison with CA-M2F.} In Mask2Former~\cite{mask2former}, they use the semantic class predictions $\mathbf{V}\in [0,1]^{N \times C}$ instead of the validity score, where $C$ is the number of classes, and it empirically yields exceptional semantic segmentation results $\mathbf{Y} \in \mathbb{R}^{H\times W\times C}$ by matrix multiplication between $\mathbf{V}$ and $\mathbf{R}$.
We explain this matrix multiplication as the calculation of the likelihood $p(x_{h,w}\in c)$, where $x_{h,w}$ and $c$ denote the pixel at location $(h, w)$ of input image $\mathbf{X}$ and the class $c \in C$: 
\begin{equation}
\label{eq1}
\max_{c}{p(x_{h,w}\in c)} = \max_{c}\sum_{n=1}^N p(\ x_{h,w} \in \mathcal{R}_n \cap x_{h,w} \in c) \nonumber
\vspace{0pt}
\end{equation}
\begin{equation}
\label{eq1.5}
\begin{split}
& = \max_{c}\sum_{n=1}^N p(x_{h,w} \in \mathcal{R}_n) \times p(x_{h,w} \in c|x_{h,w} \in \mathcal{R}_n)    \\
& = \max_{c}\sum_{n=1}^N r_{n,h,w} \times v_{n,c} = \max_{c}(\mathbf{R}^\top \cdot \mathbf{V})_{h,w,c},
\end{split}
\vspace{-0pt}
\end{equation}
where the $r_{n, h, w}$ and $v_{n, c}$ at $\mathbf{R}$ and $\mathbf{V}$ indicate the probability of pixel $x_{h,w}$ belonging to region $\mathcal{R}_n$ and region $\mathcal{R}_n$ belonging to class $c$. Given that the pixels in the same region follow the same class distribution, we have $p(\mathcal{R}_n \in c) = p(x_{h,w} \in c|x_{h,w} \in \mathcal{R}_n)$. 

CA-M2F removes the class loss and uses $\mathbf{R}$ as the region proposals, which are demonstrated to be unsatisfactory due to the redundancy of regions. To reduce useless regions, some methods keep the class prediction but reduce it to a binary classification indicating the validity of the region. And they utilize hard assignments during inference to generate the region proposals.

\begin{figure*}[htp]
  \centering
  \includegraphics[width=0.9\linewidth]{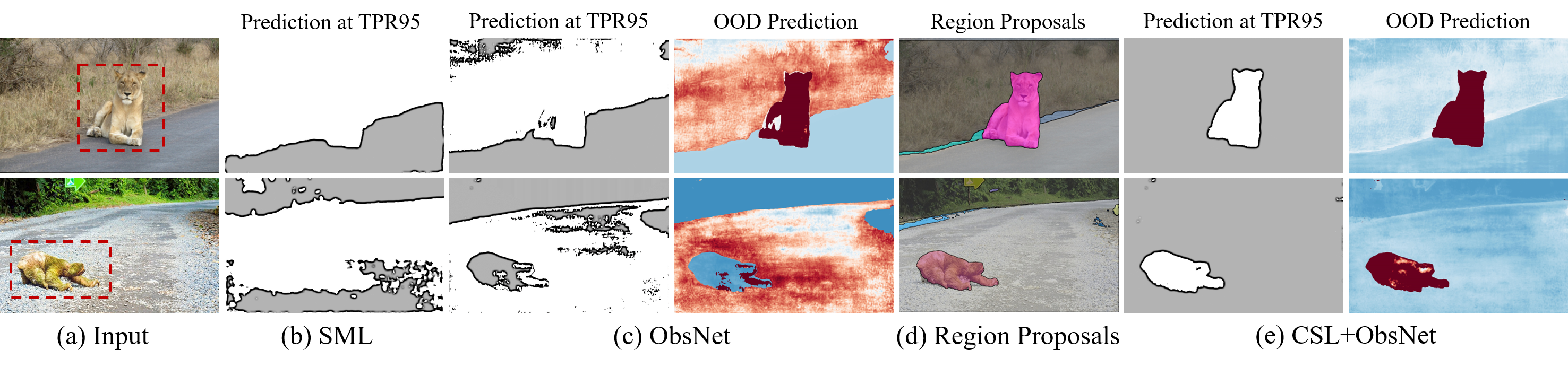}
  \vspace{0pt}
  \caption{Visualisations for hard mask predictions at TPR=95\% and per-pixel Out-of-distribution (OOD) scores. We compare the results of SML, and ObsNet with our proposed CSL. In the hard mask predictions, white and gray indicate being predicted to be OOD. In the OOD scores, the red and blue intensity values correspond to the magnitudes of the OOD scores above and below the decision boundary, respectively. (d) shows the region proposals from our CSL with scheme$_2$.
  }
  \vspace{0pt}
  \label{fig1}
\end{figure*}

\subsubsection{Soft Assignment.}
However, the hard assignment requires a manually selected threshold, which limits the generalization, and multiple experiments demonstrate its limited performance on OOD objects (Figure~\ref{fig6}). 
Thus, we creatively propose a soft assignment module
to maximize the objective of the class-agnostic semantic segmentation: $p(x_{h,w} \in \mathcal{R}_n \cap \mathcal{R}_n \in  \mathcal{V})$, where $\mathcal{R}_n \in \mathcal{V}$ denotes region $\mathcal{R}_n$ being valid.
In CSL, we use binary cross-entropy losses for optimal validity score and region score by maximizing the likelihoods of $p(x_{h,w} \in \mathcal{R}_n)$ and $p(x_{h,w} \in\mathcal{V}|x_{h,w}\in\mathcal{R}_n)$. This allows us to interpret $v_{n}$, in the validity score matrix $\mathbf{V}$,
as the probability of the pixels in the region $\mathcal{R}_n$ being valid, which we call as the region's validity score. Since multiple overlapping region prototypes exist, $\mathbf{V}$ helps select the valid masks. For instance, simple images with few objects result in fewer valid regions than complex ones. Similarly, we can interpret $r_{n,h,w}$, in the region prediction matrix $\mathbf{R}$, as the probability that pixel $x_{h,w}$ belongs to region $\mathcal{R}_n$, which we call as the region score. The objective for class-agnostic segmentation can be derived from these two likelihoods as follows:
\begin{equation}
\label{eq2}
	  \max_{n} p(x_{h,w} \in \mathcal{R}_n \cap \mathcal{R}_n \in  \mathcal{V}) \nonumber
\end{equation}
\begin{equation}
\label{eq2.5}
	\begin{split}
        & = \max_{n}  p(x_{h,w} \in \mathcal{R}_n) \times p(x_{h,w} \in \mathcal{V} \ | \ x_{h,w} \in \mathcal{R}_n)  \\
        & = \max_{n} (r_{n,h,w} \times v_n) = \max_{n} (\mathbf{R}^\top * \mathbf{V})_{h,w,n},
	\end{split}
\end{equation}
which maximizes the probability that the pixel $x_{h,w}$ is from region $\mathcal{R}_n$ and the region $\mathcal{R}_n$ is valid, 
where * denotes pixel-wise multiplication.
Note that given that pixel $x_{h,w}$ belongs to region $\mathcal{R}_n$, the validity of $\mathcal{R}_n$ can be represented by $x_{h,w}$ because the validity score is assigned per region.


\subsubsection{Comparison with hard assignment.} For panoptic segmentation, Mask2Former and EntitySeg employ hard assignment during inference
Specifically, a binary region mask is generated from each region prediction by checking at each pixel if the mask score exceeds a certain threshold. The region masks are then stacked in ascending order of validity score, where a mask with a higher validity score covers those with lower scores. However, this approach has limitations. First, hard assignment employs a fixed threshold to filter out low-score regions, which often results in missed pixels that are not allocated to any region (a, c, and d in Figure~\ref{fig6}). Second, it performs poorly in complex and detailed scenes as the final results are obtained based on hard region masks instead of per-pixel scores in the soft assignment (f in Figure~\ref{fig6}).


\subsubsection{Mask Split.}Another problem is that existing methods such as entity segmentation \cite{qi2022open} work well when training with instance-wise labels while failing with semantic-wise labels. We believe and demonstrate it's because the semantic-wise labels still contain class information, which introduces the bias of seen class. Therefore, we present a simple yet effective preprocessing method named mask split. 
Annotation of semantic segmentation often consists of multiple disconnected regions of the same class in a single mask, which forces the model to predict all pixels from the same class to the same mask, thereby enforcing the embeddings of the pixels from the same semantic class to be similar, which causes the bias. To overcome this limitation, we propose a simple yet effective method called Mask Split to overcome this limitation. 
We split these two disconnected components as depicted in Supplementary Material. This approach reduces the class information and allows the model to predict instances without being biased toward any particular class.

\begin{table}[]
\centering
\scalebox{0.9}{
\begin{tabular}{cccccc}
\bottomrule[2pt]
\multirow{2}{*}{Method} & \multirow{2}{*}{OOD} & \multicolumn{2}{c}{Pixel Level} & \multicolumn{2}{c}{Component Level} \\
                        &                      & AUPR            & FPR$_{95}$       & sIoU gt          & mean F1          \\ \hline
PEBAL \scriptsize\color{lightgray}{}         &         \twemoji{check mark}             & 49.1            & 40.8          & 38.9             & 14.5             \\
ME \scriptsize\color{lightgray}{}          &         \twemoji{check mark}             & 85.5            & 15.0          & 49.2             & 28.7             \\
DH \scriptsize\color{lightgray}{}            &         \twemoji{check mark}             & 78.0            & 9.8           & 54.2             & 31.1             \\
SynBoost \scriptsize\color{lightgray}{}       &         \twemoji{check mark}             & 56.4            & 61.9          & 34.7             & 10.0          \\
IR \scriptsize\color{lightgray}{}            &         \twemoji{multiply}             & 52.3            & 25.9          & 39.7             & 12.5             \\\hline
ObsNet \scriptsize\color{lightgray}{}        &         \twemoji{multiply}             & 75.4            & 26.7          & 44.2             & 45.1             \\
 \ttfamily +CSL$_1$             &         \twemoji{multiply}             & 79.9   & \textbf{7.1}  & 46.1    & 50.2    \\
 \ttfamily +CSL$_2$              &         \twemoji{multiply}             & \textbf{80.1}   & 7.2  & \textbf{46.5}    & \textbf{50.4}    \\
\bottomrule[2pt]
\end{tabular}}
\vspace{0pt}
\caption{Results on SMIYC-AT.} 
\vspace{0pt}
\label{table0}
\end{table}

\begin{table}[]
\centering
\scalebox{0.9}{
\begin{tabular}{cccccc}
\bottomrule[2pt]
\multirow{2}{*}{Method} & \multirow{2}{*}{OOD} & \multicolumn{2}{c}{Pixel Level} & \multicolumn{2}{c}{Component Level} \\
                        &                      & AUPR            & FPR$_{95}$      & sIoU gt          & mean F1          \\ \hline

ME \scriptsize\color{lightgray}{}                    &         \twemoji{check mark}             & 85.1            & 0.75          & 47.9             & 50.4             \\
DH \scriptsize\color{lightgray}{}                     &          \twemoji{check mark}            & 80.8            & 6.02          & \textbf{48.5}             & 55.6             \\
SynBoost \scriptsize\color{lightgray}{}               &          \twemoji{check mark}             & 71.3            & 3.15          & 44.3             & 37.6  \\           
RI \scriptsize\color{lightgray}{}                    &           \twemoji{multiply}           & 54.14           & 47.1          & \textbf{57.6}             & 36.0             \\ 
IR \scriptsize\color{lightgray}{}            &         \twemoji{multiply}             & 37.7            & 4.7         & 16.6             & 8.4             \\\hline
DaCUP \scriptsize\color{lightgray}{}                   &            \twemoji{multiply}          & 81.5            & 1.13          & 37.7             & 46.0             \\
 \ttfamily +CSL$_1$                    &            \twemoji{multiply}          & 86.8   & 0.9  & 44.3    & 50.7    \\
 \ttfamily +CSL$_2$                    &            \twemoji{multiply}          & \textbf{87.1}   & \textbf{0.7}  & 44.7    & \textbf{51.0}    \\ 
\bottomrule[2pt]
\end{tabular}}
\vspace{0pt}
\caption{Results on SMIYC-OT.}
\vspace{0pt}
\label{tableot}
\end{table}

\begin{table}[]
\centering
\scalebox{0.9}{
\begin{tabular}{cccccc}
\bottomrule[2pt]
\multirow{2}{*}{Method} & \multirow{2}{*}{OOD} & \multicolumn{2}{c}{Pixel Level} & \multicolumn{2}{c}{Component Level} \\
                        &                      & AUPR            & FPR$_{95}$       & sIoU gt          & mean F1          \\ \hline
ME \scriptsize\color{lightgray}{}                     &       \twemoji{check mark}               & 77.90           & 9.70          & 45.90            & 49.92            \\
SynBoost  \scriptsize\color{lightgray}{}              &       \twemoji{check mark}               & 81.71           & 4.64          & 36.83            & 48.72            \\ 
RI   \scriptsize\color{lightgray}{}                   &       \twemoji{multiply}               & 82.93           & 35.75         & 49.21            & 52.25            \\\hline
DaCUP   \scriptsize\color{lightgray}{}                &       \twemoji{multiply}               & 81.37           & 7.36          & 38.34            & 51.24            \\
 \ttfamily +CSL$_1$                   &         \twemoji{multiply}             & 83.07  & 6.88 & 40.43   & 51.57   \\ 
 \ttfamily +CSL$_2$                   &         \twemoji{multiply}             & 83.41  & 6.92 & 40.89   & 51.36   \\
NFlowJS               &       \twemoji{multiply}               & 89.28           & 0.65          & 54.63            & 61.75            \\
 \ttfamily +CSL$_1$                    &        \twemoji{multiply}              & 89.48  & \textbf{0.48} & 54.78   & 62.25  \\
 \ttfamily +CSL$_2$                    &        \twemoji{multiply}              & \textbf{89.79}  & 0.51 & \textbf{55.01}   & \textbf{62.37}  \\
\bottomrule[2pt]
\end{tabular}}
\vspace{0pt}
\caption{Results on LAF NoKnown.}
\vspace{0pt}
\label{table4}
\end{table}

\begin{table*}[]
\centering
\scalebox{0.9}{
\begin{tabular}{c|c|ccc|ccc|ccc}
\bottomrule[2pt]
                                     &                                                                       & \multicolumn{3}{c|}{SMIYC (AT)-val} & \multicolumn{3}{c|}{SMIYC (AT)-test} & \multicolumn{3}{c}{Road Anomaly} \\ \cline{3-11} 
\multirow{-2}{*}{Method}            & \multirow{-2}{*}{\begin{tabular}[c]{@{}c@{}}OOD \\ Data\end{tabular}} & FPR$_{95}$        & AP        & AUROC     & FPR$_{95}$              & AP               & AUROC            & FPR$_{95}$       & AP        & AUROC    \\ \hline
\cline{1-1} \cline{3-11} 
SML$^\dagger$                                 & \twemoji{multiply}                                   & 51.0      & 47.7     & 81.7     & 43.33            & 44.68            & 86.57            & 49.63     & 25.71     & 81.90    \\
{ \ttfamily +CSL$_2$} & \twemoji{multiply}                                  & 22.0$_{\uparrow 29}$      & 55.4$_{\uparrow 7.8}$     & 88.4$_{\uparrow 6.7}$     & 39.7$_{\uparrow 3.7}$            & 47.2$_{\uparrow 2.6}$            & 87.5$_{\uparrow 0.86}$            & \textbf{41.03}$_{\uparrow 8.60 }$    & 31.78$_{\uparrow 6.07}$     & 84.77$_{\uparrow 2.87}$    \\ \cline{1-1} \cline{3-11} 
IR$^\dagger$                                  & \twemoji{multiply}                                   & -          & -         & -         & 32.17            & 49.36            & 87.03            & 69.79     & 33.43     & 79.92    \\
{  \ttfamily+CSL$_2$} & \twemoji{multiply}                                   & -          & -         & -         & 21.7$_{\uparrow 10.4}$            & 54.5$_{\uparrow 5.1}$            & 89.3$_{\uparrow 2.3}$            & 59.16$_{\uparrow 10.63}$     & 35.48$_{\uparrow 2.05}$     & 82.45$_{\uparrow 2.53}$    \\ \cline{1-1} \cline{3-11} 
ObsNet$^\dagger$                               & \twemoji{multiply}                                  & 40.2      & 72.7     & 91.9     & 61.73            & 56.91            & 86.22            & 64.25     & 48.13     & 83.18    \\
{ \ttfamily +CSL$_2$} & \twemoji{multiply}                                   & 25.0$_{\uparrow 15.2}$      & 75.5$_{\uparrow 2.8}$     & 95.2$_{\uparrow 3.3}$     & 32.1$_{\uparrow 29.6}$            & 64.8$_{\uparrow 7.8}$            & 92.24$_{\uparrow 6.02}$            & 47.21$_{\uparrow 17.04}$     & 53.24$_{\uparrow 5.11}$     & 86.92$_{\uparrow 3.74}$    \\ \cline{1-1} \cline{3-11} 
ObsNet v2$^\dagger$                             & \twemoji{multiply}                                   & 30.3      & 74.5     & 93.7     & 26.69            & 75.44            & 93.80            & 55.75     & 54.64     & 86.78    \\
{ \ttfamily +CSL$_2$} & \twemoji{multiply}                                  & \textbf{5.8}$_{\uparrow 24.5}$       & \textbf{83.6}$_{\uparrow 9.1}$     & \textbf{97.4}$_{\uparrow 3.7}$     & \textbf{7.16}$_{\uparrow 19.5}$             & \textbf{80.1}$_{\uparrow 4.6}$            & \textbf{96.46}$_{\uparrow 2.66}$            & 43.80$_{\uparrow 11.95}$     & \textbf{61.38}$_{\uparrow 6.74}$     & \textbf{91.08}$_{\uparrow 4.3}$  \\
\bottomrule[2pt]
\end{tabular}}
\vspace{0pt}
\caption{Quantitative results on SMIYC (Anomaly Track) and Road Anomaly. We show the results obtained by combining CSL$_2$ with $3$ well-established OOD segmentation methods (indicated by †). The best results are highlighted in bold.}
\vspace{0pt}
\label{table 1}

\end{table*}

\subsubsection{Structure-Constrained Fusion.}
Intuitively, making predictions at each pixel independently does not benefit from the predictions at other, nearby pixels, which are correlated. To address this, we introduce structure-constrained rectification (SCF), which utilizes structure constraints, that interrelate predictions at different pixels, to optimize per-pixel predictions. This helps improve performance on multiple tasks.  
Our proposed approach leverages soft assignment to generate region proposals $\mathcal{R} \in \{0,1\}^{H\times W\times N}$. These proposals, along with the per-pixel distribution $\mathbf{D} \in \mathbb{R}^{H \times W \times C}$ from scheme $1$ or scheme $2$, are fed into our proposed SCF. For OOD segmentation, we set $C$ to 1 since the prediction is binary, and we only need to consider the probability of belonging to OOD. And, for the domain adaptation (DA) and zero-shot semantic segmentation (ZS3) tasks, $C$ is equal to the number of classes.

Each $\mathbf{D}_c \in \mathbb{R}^{H\times W}$ indicates per-pixel distribution for class $c$, where $c \in \{1,...,C\}$. We compute the region-wise score as the average of the pixel-wise scores within each region proposal $\mathcal{R}_n \in \{0,1\}^{H\times W}$, where $n\in\{1,...,N\}$. Then we combine the region-wise scores and the pixel-wise scores to obtain the hybrid score $\mathbf{H}$ using the equation:
\begin{equation} 
\label{eq6}
        \mathbf{H}_{c, n} = \frac{\sum_{h,w} \mathbf{D}_{c, n} * \mathcal{R}_n}{\sum_{h,w} \mathcal{R}_n} \times \mathbf{D}_{c, n}, \ \ n \in \{1,...,N\}
\end{equation}
where $\mathbf{H}_{c, n}$, $\mathbf{D}_{c, n}$ indicates the hybrid score and per-pixel distribution for class $c$ within region $n$, $*$ is the pixel-wise multiplication and $\triangle$ indicates pixel-wise multiplication.

\begin{table*}[]
\centering
\scalebox{0.9}{
\begin{tabular}{cccccc|ccc}
\bottomrule[2pt]
\multirow{2}{*}{Method} & \multirow{2}{*}{ST} & \multirow{2}{*}{RT} & \multicolumn{3}{c|}{COCO-stuff} & \multicolumn{3}{c}{PASCAL VOC 2012} \\
                        &                     &                     & mIoU(S)    & mIoU(U)   & hIoU   & mIoU(S)     & mIoU(U)    & hIoU     \\ \hline
ZegFormer               &         \twemoji{multiply}            &        \twemoji{multiply}             & 36.6       & 33.2      & 34.8   & 86.4        & 63.6       & 73.3     \\
 \ttfamily +CSL$_2$                   &          \twemoji{multiply}           &         \twemoji{multiply}            & 37.5 $_{\uparrow 0.9}$      & 36.2 $_{\uparrow 3}$     & 36.9 $_{\uparrow 2.1}$  & 87.1  $_{\uparrow 0.7}$      & 68.6  $_{\uparrow 5}$     & 76.9 $_{\uparrow 3.6}$   \\ \hline
 ZSSeg                   &         \twemoji{multiply}            &        \twemoji{multiply}             & 39.3       & 36.3      & 37.8   & 83.5        & 72.5       & 77.5     \\
  \ttfamily +CSL$_2$                   &         \twemoji{multiply}            &        \twemoji{multiply}             & 40.1 $_{\uparrow 0.8}$      & 38.3  $_{\uparrow 2}$    & 39.2 $_{\uparrow 1.4}$  & 84.7  $_{\uparrow 1.2}$      & 76.9  $_{\uparrow 4.4}$     & 80.6  $_{\uparrow 3.1}$   \\ \hline
 ZegCLIP                 &         \twemoji{multiply}            &         \twemoji{multiply}            & 40.2       & 41.4      & 40.8   & 91.9        & 77.8       & 84.3     \\
  \ttfamily +CSL$_2^*$                   &         \twemoji{multiply}            &         \twemoji{multiply}            & 40.4  $_{\uparrow 0.2}$     & 42.8 $_{\uparrow 1.4}$     & 41.6 $_{\uparrow 0.8}$  & 92.3 $_{\uparrow 0.4}$       & 79.4 $_{\uparrow 1.6}$      & 85.5 $_{\uparrow 1.2}$ 
\\ \bottomrule[2pt]  
\end{tabular}
}
\vspace{0pt}
\caption{Quantitative results for ZS3 on COCO-stuff and PASCAL VOC benchmarks. The “mIoU(S)”, “mIoU(U)”, and “hIoU” denote the mIoU of seen classes, unseen classes, and their harmonic mean. "ST" and "RT" denote self-training and re-training.}
\vspace{0pt}
\label{table5}
\end{table*}

\section{Experimental Results}
\subsection{Experimental Setup}
In all our experiments \footnote{Except for experiments marked with $*$, which uses ResNet100 as the backbone.}, we utilize ResNet50 as the backbone and FPN as the pixel decoder. All experiments for OOD segmentation are performed without any OOD data. In the DA and ZS3 experiments, we use the same training data as the comparative methods. CSL$_1$ and CSL$_2$ represent scheme 1 and 2, and CSL$_2$ doesn't require retraining. Additional details and results for the benchmarks and implementation can be found in the supplementary material, and we plan to make the source code publicly available.

\begin{table*}[]
\centering
\scalebox{0.9}{
\begin{tabular}{c|ccccccccccccccccccc|c}
\bottomrule[2pt]
Method      & \rotatebox{90}{road}  & \rotatebox{90}{sidewalk} & \rotatebox{90}{building} & \rotatebox{90}{wall}  & \rotatebox{90}{fence} & \rotatebox{90}{pole}  & \rotatebox{90}{light} & \rotatebox{90}{sign}  & \rotatebox{90}{veg} & \rotatebox{90}{terrain} & \rotatebox{90}{sky}   & \rotatebox{90}{person} & \rotatebox{90}{rider} & \rotatebox{90}{car}   & \rotatebox{90}{truck} & \rotatebox{90}{bus}   & \rotatebox{90}{train} & \rotatebox{90}{motor} & \rotatebox{90}{bicycle} & mIoU  \\ \hline
Source Only & 79 & 39    & 75    & 26 & 25 & 34 & 34 & 39 & 82      & 18   & 84 & 58  & 37 & 70 & 19 & 15 & 5  & 22     & 54   & 43  \\
 \ttfamily +CSL$_2$        & \textbf{82} & 36    & \textbf{78}    & \textbf{28} & \textbf{29} & \textbf{40} & \textbf{45} & \textbf{48} & \textbf{83}      & \textbf{25}   & 81 & \textbf{68}  & \textbf{45} & \textbf{81} & \textbf{24} & \textbf{20} & \textbf{7}  & \textbf{24}     & \textbf{57}   & \textbf{47}$_{\uparrow 4}$ \\ \hline
AdvEent         & 94 & 59    & 85    & 28 & 26 & 38 & 43 & 43 & 86      & 28   & 89 & 61  & 36 & 87 & 32 & 46 & 25 & 25     & 57   & 52 \\
 \ttfamily +CSL$_2$        & \textbf{94}    & \textbf{60}    & \textbf{85}    & 28  & \textbf{35} & \textbf{45} & \textbf{48} & \textbf{50} & \textbf{86}      & 28   & \textbf{89} & \textbf{65}  & \textbf{46} & \textbf{87} & \textbf{38}  & \textbf{49} & \textbf{32} & 24      & \textbf{59}   & \textbf{56}$_{\uparrow 4}$ \\ \hline
DAFormer         & 96 & 73    & 89    & 40 & 44 & 49 & 53 & 60 & 58      & 49   & 91 & 71  & 45 & 91 & 75 & 77 & 64 & 55     & 61   & 65 \\
 \ttfamily +CSL$_2^*$       & \textbf{96}    & \textbf{74}    & \textbf{90}    & \textbf{51}  & \textbf{48} & \textbf{52} & \textbf{56} & \textbf{65} & \textbf{89}      & \textbf{48}   & \textbf{91} & \textbf{76}  & \textbf{45} & \textbf{93} & \textbf{77}  & \textbf{80} & \textbf{68} & \textbf{56}      & \textbf{66}   & \textbf{70}$_{\uparrow 4}$ \\
\bottomrule[2pt]
\end{tabular}}
\vspace{0pt}
\caption{Quantitative results for domain adaptation on the Synscapes2Cityscapes benchmark, where the source domain is the synthetic city scenes dataset (Synscapes) and the target domain is a real-world city scenes dataset (Cityscapes).}
\vspace{0pt}
\label{table3}
\end{table*}

\subsection{Out-Of-Distribution Segmentation}

 In the context of OOD Segmentation, Cityscapes~\cite{cityscapes} including 19 seen classes are used as the training sets, while OOD images containing other classes beyond the seen classes are utilized for testing purposes. Several approaches leverage OOD images with ground truth labels from larger datasets to enrich the training set, which overlaps with the OOD classes in the test set. Thus, to ensure fairness, all methods are differentiated based on the usage of OOD data and our proposed CSL is free of OOD data.
\vspace{-0pt}
\subsubsection{Comparison with SOTA Methods}
\vspace{-0pt}
Table~\ref{table0}-\ref{table 1} show our results compared with existing methods on the SMIYC~\cite{SMIYC} Anomaly Track, Obstacle Track, LostAndFound-NoKnow \cite{laf}, and Road Anomaly \cite{imageresynthesis}. 
There are $5$ metrics for evaluation: (a) pixel-wise area under the precision-recall curve (AUPR), (b) pixel-wise false positive rate at a true positive rate of 95\% (FPR$_{95}$), (c) adjusted Intersection over Union averaged over all ground truth segmentation components (sIoU gt), (d) component-wise F1-score averaged over different detection thresholds (mean F1), and (f) area under the receiver operating characteristics (AUROC). 
\textbf{SMIYC (Anomaly Track)} consists of real-world images, where each image may contain multiple OOD samples of different sizes from various categories. In SMIYC (AT), our proposed approach CSL outperforms all methods without OOD data by a substantial margin based on ObsNet, \textit{e.g.}, CSL surpasses the former state of art method ObsNet \cite{obsnet} by $4.7\%$, $19.5\%$, $2.3\%$, and $5.3\%$ in AUPR, FPR$_{95}$, sIou gt, and mean F$1$. CSL even reaches state-of-the-art performance in terms of FPR$_{95}$ and mean F$1$ across all methods including those leverage the OOD data. 
\textbf{SMIYC (Obstacle Track)} focuses on evaluating the ability to detect small-size obstacles on the road. In SMIYC (OT), CSL improves the former approach DaCUP \cite{dacup} by $5.6\%$, $0.43\%$, $7\%$, and $5\%$ in AUPR, FPR$_{95}$, sIou gt, and mean F$1$ and achieve the state of art among all approaches in AUPR, FPR$_{95}$, and mean F$1$.  
\textbf{LostAndFound NoKnown} also focuses on evaluating the ability to detect small-size obstacles on the road and CSL improves the former approach DaCUP \cite{dacup} by $2.04\%$, $0.44\%$, $2.55\%$, and $0.12\%$ in AUPR, FPR$_{95}$, sIou gt, and mean F$1$. And achieves state-of-the-art performance when combined with NFlowJS \cite{nflowjs}.
\textbf{Road Anomaly} has a similar setting with SMIYC (AT). As shown in Table~\ref{table 1}, CSL achieves state of art performance among all methods including those with OOD data by improving the performance of ObsNet \cite{obsnet} by $6.74\%$, $11.59\%$, and $4.3\%$ in AP, FPR$_{95}$, and AUROC.

\subsubsection{Combination with Existing Methods without Retraining}

We combine CSL with three existing OOD segmentation methods (SML, Image Resynthesis, and ObsNet) with scheme $2$, which doesn't require retraining, and compared their performance in Table~\ref{table 1}. Noted that the performance of ObsNet is affected by the input size. Therefore, we use the ObsNet and ObsNet v2 to represent the experiments we use the original image size and fixed-smaller image size, \textit{i.e.}, $512 \times 1024$. CSL outperformed all other methods in both benchmarks and even surpassed methods that use OOD data. Some methods achieved a decent AP but a poor FPR$_{95}$ due to the difficulty of extracting OOD samples. ObsNet and ObsNet v2 achieved a high FPR$_{95}$ in SMIYC (AT)-test, but our CSL significantly reduced it by 29.57 and 22.57, respectively.
Figure \ref{fig1} visually compares our proposed CSL with existing OOD segmentation methods, where we use ObsNet v2 to represent ObsNet due to its better performance. SML struggles to get acceptable results, and ObsNet produces decent AP but fails to achieve high recall with low FPR as shown in (e). In contrast, CSL demonstrates robustness to OOD samples as shown in (e).

\subsection{Zero-Shot Semantic Segmentation}
Table~\ref{table5} presents a comparison of our proposed CSL method with previous state-of-the-art zero-shot semantic segmentation methods. 
We adopt the scheme$_2$ to integrate CSL with existing methods, primarily due to its reduced computational cost. (Section scheme$_1$ vs scheme$_2$).
CSL outperforms ZegFormer, ZSSeg~\cite{zsseg}, and ZegCLIP by $0.9\%, 0.8\%, 0.2\%$ in seen classes, $3\%, 2\%, 1.4\%$ in unseen classes, and $2.1\%, 1.4\%, 0.8\%$ in harmonic classes in COCO-stuff benchmark and outperform those 3 methods by $5\%, 4.4\%, 1.6\%$ in unseen classes in PASCAL VOC 2012 benchmark. The experiment follows the same setting as ZegFormer, using $156$ classes for training, and testing on all $171$ classes from the COCO-stuff dataset.

\subsection{Domain Adaptation in Semantic Segmentation}
The CSL approach demonstrates superior performance not only on out-of-distribution (OOD) samples but also on in-distribution (ID) samples with domain gaps. Notably, our method achieves excellent results on the Synscapes2Cityscapes benchmark, as reported in Table~\ref{table3}. In these experiments, we use Synscapes, a synthetic city scene dataset, as the source domain, and Cityscapes, a real-world city scene dataset, as the target domain. And we also choose scheme$_2$ to integrate CSL with existing methods. CSL boost source-only by $4.02\%$, AdvEnt \cite{advent} by $3.5\%$, and DAFormer~\cite{daformer} by $4.1\%$.

\subsection{Ablations}
\label{CA}

\subsubsection{Negative Impact from Class Information}

Traditional methods for semantic segmentation assign each pixel from input images to prior semantic classes. However, this approach cannot handle OOD samples. The CA-RPG method assigns each pixel to $N$ class-agnostic region prototypes, which learn more fundamental features that can represent both ID and OOD samples.
Mask2Former and Zegformer also use a query-based framework, but introducing class supervision destroys the ability for OOD segmentation. The classification loss of ID classes causes the region prototypes to distribute within the subspace of ID classes, which makes it difficult to represent OOD classes effectively.
In Figure~\ref{fig7}, we can see the results of using the None-CA approach versus the CA-training approach on an image of a skier. The embeddings for the skier and background are not easily separable using the None-CA approach, while the CA-training approach allows for a better representation of both classes.

\begin{figure}[htp]
  \centering
  \includegraphics[width=1\linewidth]{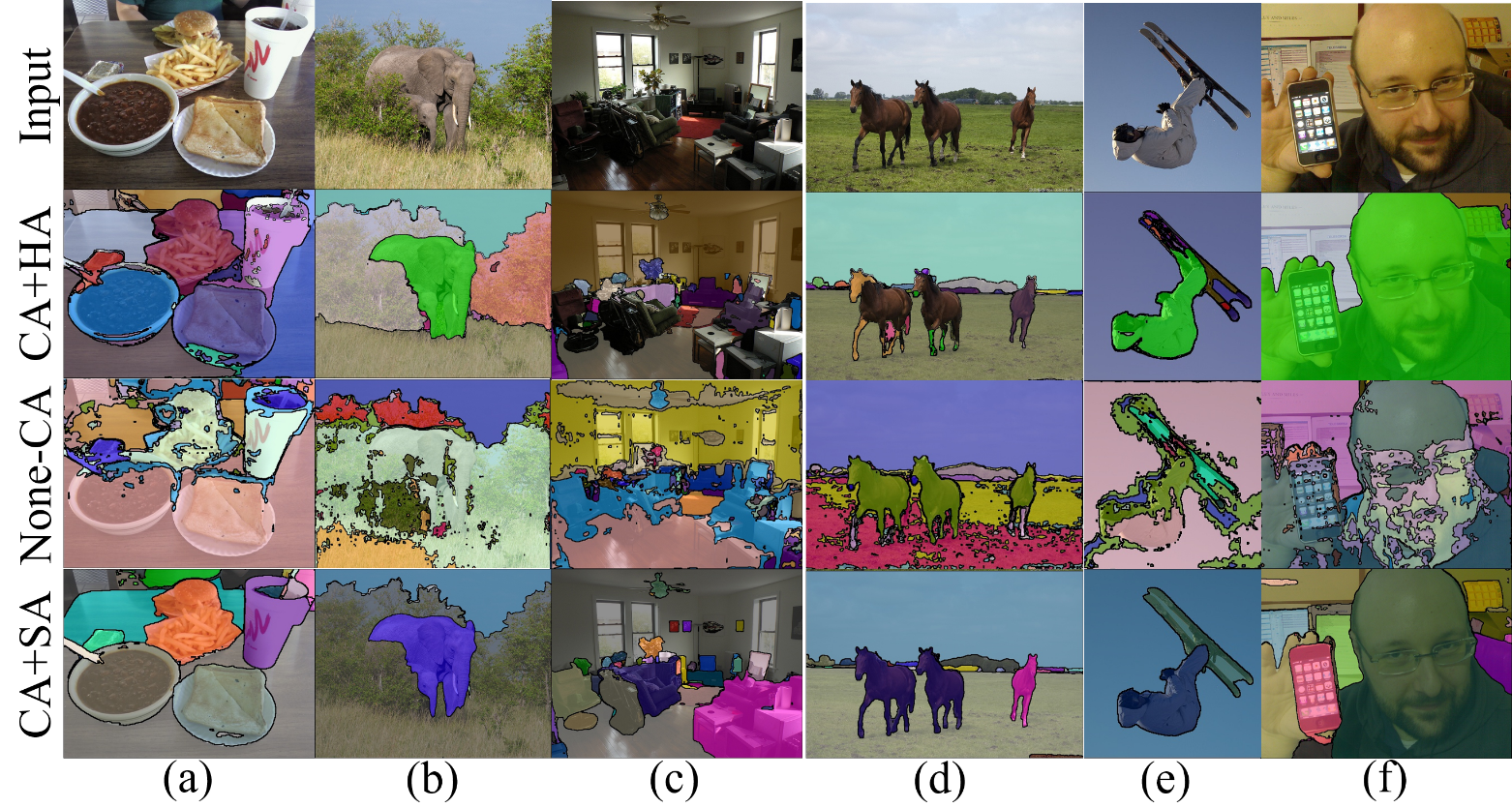}
  \vspace{0pt}
  \caption{Visualisations of the efficacy of CA-training and soft assignment. CA+HA represents CA-M2F, where the model is trained in a class-agnostic way and inferences via the hard assignment, None-CA represents the model is trained with class loss and inferences via the soft assignment, and CA+SA represents CSL, where the model trained in a class-agnostic way and inferences via the soft assignment. Note that the model is only trained on the Cityscapes and tested on COCO-stuff.
  }
  \vspace{0pt}
  \label{fig6}
\end{figure}

\vspace{-0pt}
\subsubsection{CA Training and Soft Assignment}
\vspace{-0pt}

Quantitative results in Table~\ref{table2} demonstrate the effectiveness of CA training and our proposed soft assignment (SA). Before evaluation, we count the ground truth labels corresponding to all pixels in each region proposal and select the label with the highest frequency as the class of the entire region. This post-processing method is proposed in SMIYC~\cite{SMIYC} and allows us to use the same evaluation criteria (mIoU, fwIoU, mACC, and pACC) as semantic segmentation to assess the quality of region proposals.

We present a comparison of three approaches for training a region proposal generator: None-CA, which employs binary mask and classification loss; CA+HA, which employs CA training and hard assignment during inference; and CA+SA, our proposed approach which combines CA training with soft assignment during inference. The model is trained on Cityscapes and tested on COCO-stuff, where ID represents seen classes from Cityscapes, and OOD represents those in the COCO but not in Cityscapes. Results in Table~\ref{table2} demonstrate that both CA training and soft assignment significantly improve performance across all metrics. Figure~\ref{fig6} visually illustrates the improvement. None-CA fails on most unseen objects, while CA+HA produces decent results but struggles with challenging cases such as indoor scenes, multiple animals, and small accessories. Soft assignment overcomes the limitations of hard assignment by assigning regions pixel-wise, providing more refined segmentation results.

\begin{table}[]
\centering
\scalebox{0.9}{
\begin{tabular}{cccc}
\bottomrule[2pt]
            & None-CA & CA+HA   & CA+SA \\ \hline
mIoU        & 45.99 & 49.17$_{\uparrow 3.18}$ & 51.34$_{\uparrow 5.35}$ \\
ID-mIoU   & 68.44   & 76.35$_{\uparrow 7.91}$   & 77.20$_{\uparrow 8.76}$ \\
OOD-mIoU & 43.83   & 46.56$_{\uparrow 2.73}$   & 48.85$_{\uparrow 5.02}$ \\
mACC        & 58.46   & 60.31$_{\uparrow 1.85}$   & 63.10$_{\uparrow 4.64}$ \\
\bottomrule[2pt]
\end{tabular}}
\vspace{0pt}
\caption{Ablation study of CA-training and Soft Assignment. The model is trained on the Cityscapes-train and tested on the COCO-stuff.}
\vspace{0pt}
\label{table2}
\end{table}

\begin{figure}
  \centering
  \includegraphics[width=0.78\linewidth]{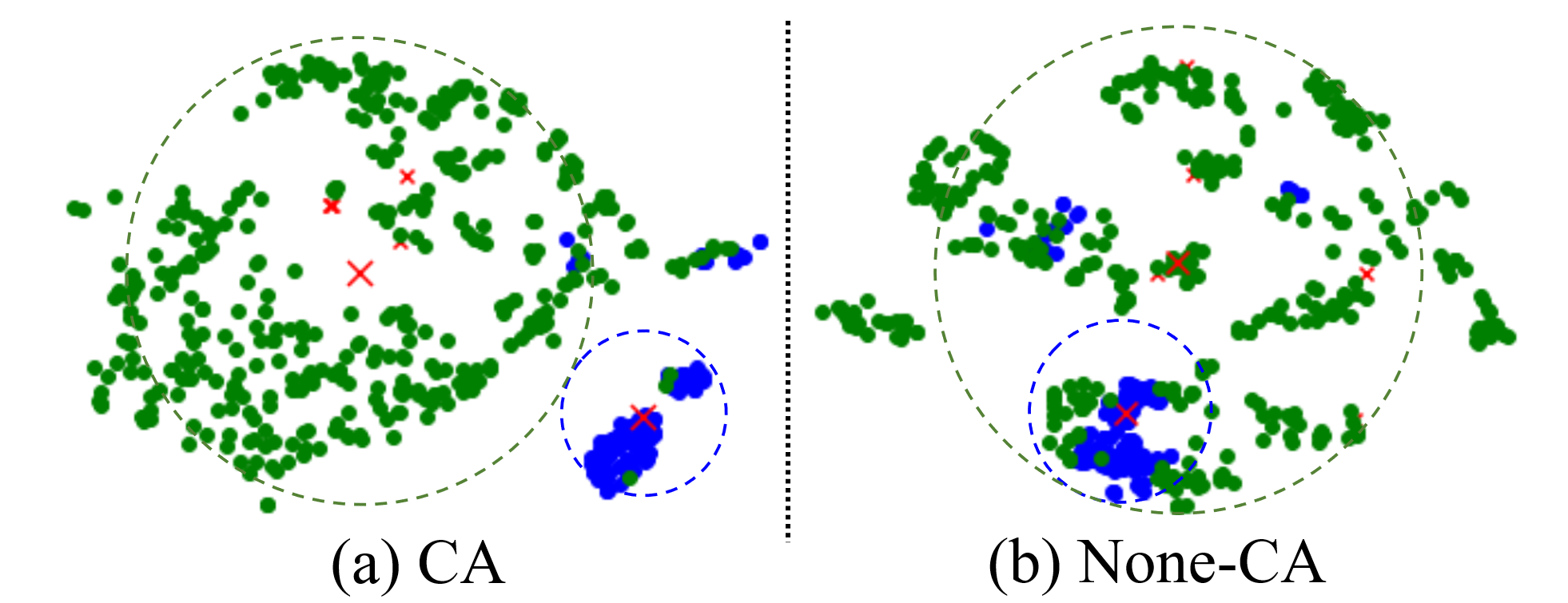}
  \vspace{0pt}
  \caption{Embedding visualisations of Figure \ref{fig6}-(e) by T-SEN. We plot the region prototypes as red times symbols, the per-pixel embeddings from the background as green bullets, and the skier as blue bullets. The sizes of the prototypes indicate the validity scores.
  }
  \vspace{0pt}
  \label{fig7}
\end{figure}

\subsubsection{Comparison with Segment Anything Model}
A notable contribution of CSL is its capability to segment out-of-distribution (OOD) objects without relying on any OOD data, utilizing only minimal training data. In this section, we employ a foundation model, SAM~\cite{sam}, which is used by many recent works~\cite{opennerf}, to produce CA region proposals and subsequently integrate it with ObeNet~\cite{obsnet}, DACUP~\cite{dacup}, and NFlowJS~\cite{nflowjs} in the SMIYC-AT, OT, and LAF NoKnown benchmarks. This approach yields an improvement of $1.7\%$ in AUPR and $0.3\%$ in FPR$_{95}$ on average across those three benchmarks compared with CSL in Table \ref{table0}, \ref{tableot}, and \ref{table4}, which demonstrates that the quality of CA region proposals generated by CSL is satisfied, even in the absence of any OOD data. We believe the constraining factor influencing the outcome appears to be the classification accuracy of each region, rather than segmentation quality. More results are shown in the Appendix.

\subsection{Scheme$_1$ vs Scheme$_2$} In Tables~\ref{table0}, \ref{tableot}, and \ref{table4}, we present results for scheme 1 and 2. While scheme$_1$ trails by approximately $0.3\%$ in AUPR relative to scheme$_2$, the results on FPR$_{95}$ display a mix of advantages for both methods. Notably, scheme$_2$ demonstrates efficiency in training, requiring half the iterations to match the performance of scheme$_1$. For context, in our integration experiments with ZegCLIP~\cite{zegclip} on the COCO-stuff benchmark, scheme$_1$ demanded around 50K iterations to achieve satisfactory results, whereas scheme$_2$ reached similar benchmarks in just 25K iterations. However, another key consideration is the inference time. Scheme$_1$, being an end-to-end solution, is more efficient during inference: in our evaluations, scheme$_2$ took $33\%$ longer on average across all conducted experiments."

\section{Conclusion}
This paper presents the Class-Agnostic Structure-Constrained Learning (CSL) method for addressing the challenge of segmenting the unseen. CSL provides 2 different schemes, which can be utilized as an end-to-end framework or integrated with existing methods without retraining. Our experimental results demonstrate that CSL outperforms existing state-of-the-art methods across 3 challenging tasks.
Moreover, we have provided an analysis of the reasons behind the effectiveness of our proposed method. We believe that the ability of CSL to learn about classes not seen during training, by eliciting class-agnostic information from the ID images, is a crucial factor contributing to its superior performance.
Overall, CSL provides a promising solution for segmenting the unseen, and we hope our work will lead to other related work in this area.

\section{Acknowledgements}
The support of the Office of Naval Research under grant N00014-20-1-2444 and of USDA National Institute of Food and Agriculture under grant 2020-67021-32799/1024178 are gratefully acknowledged.

\clearpage
\bibliography{aaai24}

\end{document}